\documentclass{mva_style}
\usepackage{graphicx}
\usepackage{cite}

\usepackage{color}

\finalcopy 

\def\eg{\emph{e.g}.} 
\def\para#1{\smallskip\noindent{\bf{#1}}}

\def\cf{\emph{c.f}.~} 
 
\def\etal{\emph{et al}.~}

\newcommand{\thisheight}{\linewidth}

\begin{document}
\title{Video-Based Camera Localization Using Anchor View Detection and Recursive 3D Reconstruction}


\author{
  Hajime Taira$^{1,\dag}$
  \quad
  Koki Onbe$^{1,\dag}$
  \quad
  Naoyuki Miyashita$^{2,\ddag}$
  \quad
  Masatoshi Okutomi$^{1,\dag}$
  \\[3pt]
  $^{1}$Tokyo Institute of Technology \quad 
  $^{2}$Olympus Corporation
  \\[3pt]
  $^{\dag}${\tt \{htaira,konbe,mxo\}@ok.sc.e.titech.ac.jp} \quad
  $^{\ddag}${\tt naoyuki.miyashita@olympus.com}
}

\maketitle

\section*{\centering Abstract}
\textit{
In this paper we introduce a new camera localization strategy designed for image sequences captured in challenging industrial situations such as industrial parts inspection. To deal with peculiar appearances that hurt standard 3D reconstruction pipeline, we exploit pre-knowledge of the scene by selecting key frames in the sequence (called as anchors) which are roughly connected to a certain location. Our method then seek the location of each frame in time-order, while recursively updating an augmented 3D model which can provide current camera location and surrounding 3D structure. 
In an experiment on a practical industrial situation, our method can localize over 99\% frames in the input sequence, whereas standard localization methods fail to reconstruct a complete camera trajectory. 
}

\section{Introduction}
\noindent
Determining a location of the camera is one of the fundamental task in computer vision, supporting a growing need of 3D reconstruction such as Structure from Motion (SfM) and Simultaneous Localization and Mapping (SLAM)~\cite{Wu133DV,Schonberger-CVPR16,cadena2016past,orbslam}. 
They are also directly applicable for a video-based navigation system that suggests the temporal user location from a sequential image series. 

Such navigation is especially beneficial for some industrial robotics scenarios, 
where an operator often cannot directly observe the scene~\cite{hansenstereo,kagami20203d}. 
In specific, for an industrial parts inspection, a thin diameter probe (industrial borescope) is inserted to the inner of a product and inspects its damages or defects. 
Whereas other auxiliary sensors~\cite{esquivel2009jprs,peter2015ijrr} often are not available for practical borescopes, a pure vision-based localization can still be helpful to guess defective locations while associating to their appearances from an image sensor~\cite{kagami20203d}. 
However, common image-based camera localization techniques~\cite{Schonberger-CVPR16,sattler2017large,torii2019large}, which simultaneously estimate the camera location and surrounding 3D structure, 
often fail to reconstruct a valid model due to peculiar appearances in industrial situations. 

In this paper, we attempt to handle such challenging industrial parts inspection scenarios. 
First, to deal with special appearance in the industrial scene, we employ the 3D structure-based approach~\cite{sattler2017efficient,sattler2017large,torii2019large,taira2018inloc,taira2019inlocpami,toft2020longterm} for localization. 
Instead of reconstructing cameras for all related images at once, we efficiently localize input images by preparing a pre-constructed 3D model presenting the targeted scene and registering new images to the model. 
Second, to stably localize all of video frames captured during inspection, we design our system to register cameras in time-order while incrementally updating 3D model, which makes it easier to find locations of consecutive frames. 
Also, we employ a new technique based on a typical key-frame (called as {\it anchor}) in the input sequence that is connected to a certain object in the target scene and contributes to a stable image registration. 
We finally test our system in one specific inspection scenario for an industrial product and validates its performance in the challenging situation. 
%

\para{Related works}. 
For the input of image series, SfM~\cite{Wu133DV,Schonberger-CVPR16} and SLAM~\cite{engel2014lsd,cadena2016past,orbslam,sumikura2019openvslam} are the well known techniques to reconstruct their 6-dimensional camera poses together with surrounding 3D structure. 
Whereas most SLAM methods assume a sequential images input and obtain a camera trajectory in time-order, SfM in a recursive manner~\cite{nyberg2006recursive,bronte2014real,song2014robust,resch2015scalable} has also been developed to provide a temporal 3D reconstruction in a real-time processing. 
Since they originally obtain a scale-indeterminate 3D model, several works also estimate cameras in a real-scale by using auxiliary sensors~\cite{esquivel2009jprs,peter2015ijrr,zhang2018scale,sumikura2018scale,jiang2021vio} or adjusting the model with a known property of the scene~\cite{sattler2017large,torii2019large,kagami20203d}. 
When a pre-constructed 3D model of the targeted scene is also available, camera poses can be further accurately estimated by registering images directly to the model~\cite{sattler2017efficient,sattler2017large,torii2019large,taira2018inloc,taira2019inlocpami}.
Kroeger \etal~\cite{kroeger2014video} proposed a video registration scheme to an SfM model. Our method also registers image sequences to an SfM model while incorporating a known property of the scene to deal with challenging industrial scenario. 
%
\begin{figure*}[t]
    \centering
    \includegraphics[width=0.88\linewidth]{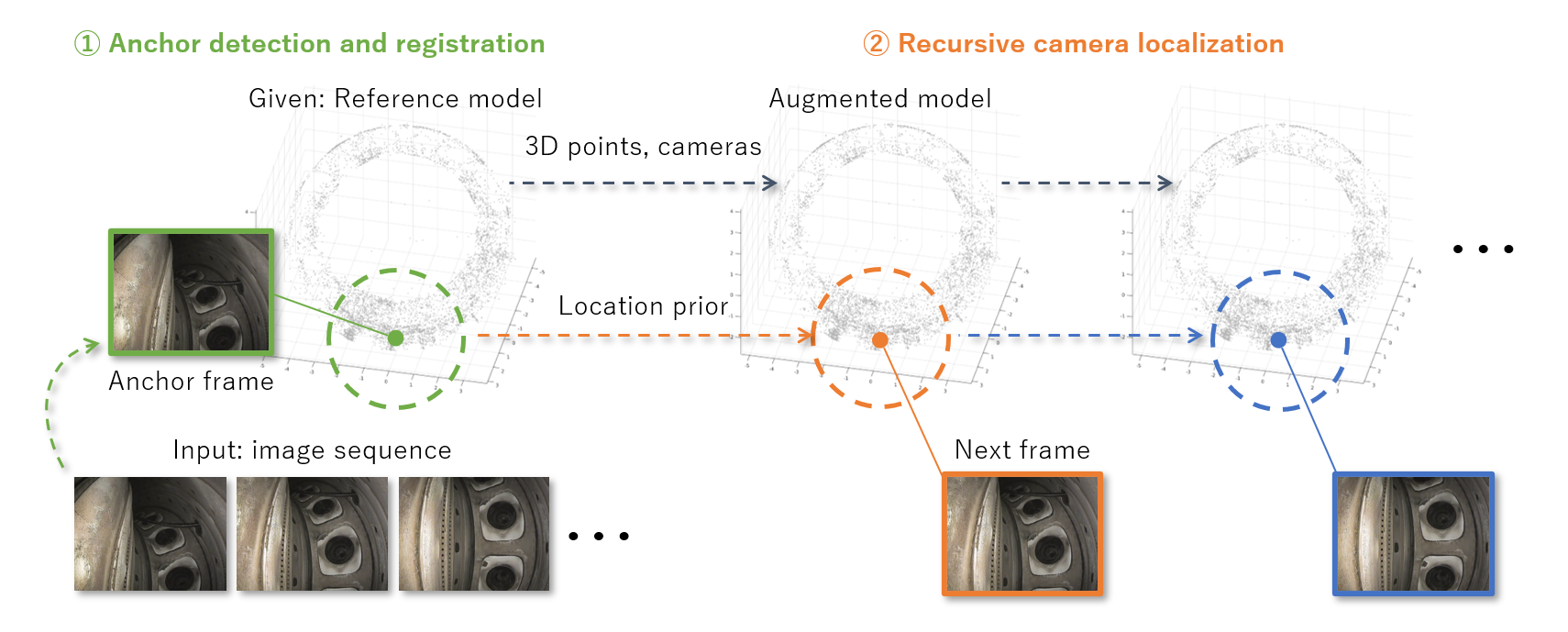}
    \caption{The whole camera localization pipeline for a sequential image series. 
    We start to reconstruct cameras from one specific anchor frame capturing the characteristic location in the targeted scene. 
    Then we localize remaining frames by incrementally updating an augmented model while registering images neighbor to the previously reconstructed camera. \label{fig:flowchart}}
\end{figure*}
\section{Video frame localization via anchor view detection and recursive 3D reconstruction \label{sec:method}}
\noindent
Fig.~\ref{fig:flowchart} shows the proposed camera localization pipeline for sequential images. 
We assume a pre-mapped SfM model consisting of 3D scene points and pre-captured (database) cameras as the reference information for the targeted scene (reference model). 
For the input of an image sequence consisting of consecutive image frames, we first find a key frame connected to a specific location in the reference model (Sec.~\ref{subsec:anchor}).  
Next we sequentially localize remaining images via an SfM-like reconstruction scheme (Sec.~\ref{subsec:recon}) which recursively update a 3D model by registering new images. 
The reconstruction will continue until all input images have been registered. 
%
\subsection{Anchor-based camera localization \label{subsec:anchor}}
\noindent
The major failure cases of reconstruction in industrial scenarios are often be attributed to the failure of localization due to the lack of texture, or highly frequent objects such as standardized industrial parts. 
To deal with these challenging appearance, we attempt to first register some key frames (anchors) which locations can be roughly determined by the pre-knowledge of the targeted scene, \eg, capturing an unique object or a marker in the scene. 
In the later section (Sec.~\ref{sec:exp}) we will describe our CNN-based anchor detector trained for one specific situation of industrial parts inspection, which constructs a subset of images potentially to be anchors out of the input sequence. 
Please note that in more general cases, such anchor frames can either be specified manually, or automatically detected via any object recognizer. 
The detected anchors can be relatively easily registered to the reference model, and also can be used as a spatial guide of other frames. 
As described in the later section (Sec.~\ref{subsec:recon}), this anchor-based approach is particularly  beneficial to stably localize sequential images.  



\para{Anchor registration to the reference model}. We register the detected anchor images to the reference model via a standard SfM scheme constructing a temporal augmented model including anchors. 
We first seek a subset of database images that share views with anchors, and perform pairwise local feature matching towards each of anchor images~\cite{lowe2004distinctive}. 
Consequent PnP and point triangulation steps~\cite{Schonberger-CVPR16} obtain the initial camera pose of anchors and 3D scene points corresponds to the local features in these newly registered images. 
The camera poses of anchors are finally refined by the bundle adjustment~\cite{Schonberger-CVPR16} which minimizes the reprojection errors of 3D scene points with respect to their local observations in the images. 
Please note that in this step we freeze parameters of database cameras and existing scene points in the reference model, so this refinement does not affect the consistency of the augmented model and referenced SfM model presenting the map information of the scene.
\subsection{Recursive 3D reconstruction for sequential images \label{subsec:recon}}
\noindent
After registering anchor frames to the SfM model, we start to determine the location of all images by incrementally register each of frames, while updating the augmented model. 

\para{Feature matching towards spatial and temporal neighbors}. The reconstruction begins from the consecutive frame of the earliest anchor frame, and continue in the time-stamp order. 
First we perform a {\it spatially-guided} feature matching that seek correspondences between local observations in the current frame and cameras in the current augmented model. 
Instead of the location knowledge used in Sec.~\ref{subsec:anchor}, this time we exploit the most recently registered camera of the input sequence as a more precise location prior. 
As in anchor registration, we collect a set of database images that are spatially neighbor to the recent camera and perform pairwise local feature matching towards the current frame. 
Additionally, we select 20 best database images via image retrieval~\cite{Schoenberger2016ACCV} and also match them towards the current frame. 

To increase the number of local connectivity of the current frame towards the augmented model, we also perform a {\it temporally-guided} feature matching that matches the current frame towards the neighbor frames, which have highly relevant appearances thus are easy to match. 
We make a set of $N$ neighbor frames that have previously been registered to the model and match them towards the current frame. In our experiments, we set $N$ as $25$ frames, which can remain a precise reconstruction with a controllable cost, but please be sure that this setting should be determined for each specific scene while also considering the video frequency. 

\para{Recursive camera localization for image sequence}. From the set of local feature matches, we can also extract correspondences between local features and the existing 3D scene points. 
If the sufficient number of these 2D-to-3D matches exists, the 6D camera pose of the current frame can be obtained by solving the Perspective-n-Points problem~\cite{fischler1981random,quan1999linear}. 
Then we triangulate new 3D scene points seen from the current frame, which can also help to register the next frame. 

We finally perform a bundle adjustment~\cite{ceres-solver} to refine the newly added scene points and cameras. 
As in anchor registration, we freeze 3D points and cameras existing in the reference model, and refine cameras for all input frames and new 3D points while minimizing the reprojection errors of the model. 
For efficiency, we perform this refinement every $10$ new frames have been registered. 
After all, we get a new temporal SfM model that provides camera location of the recent input frames, which is used to register next consecutive frames. 
The system continues to register new frames until the end of the input video, and finally obtains an SfM model including 6D cameras of all input frames. 
\begin{figure}[t]
    \centering
    \includegraphics[width=0.8\linewidth]{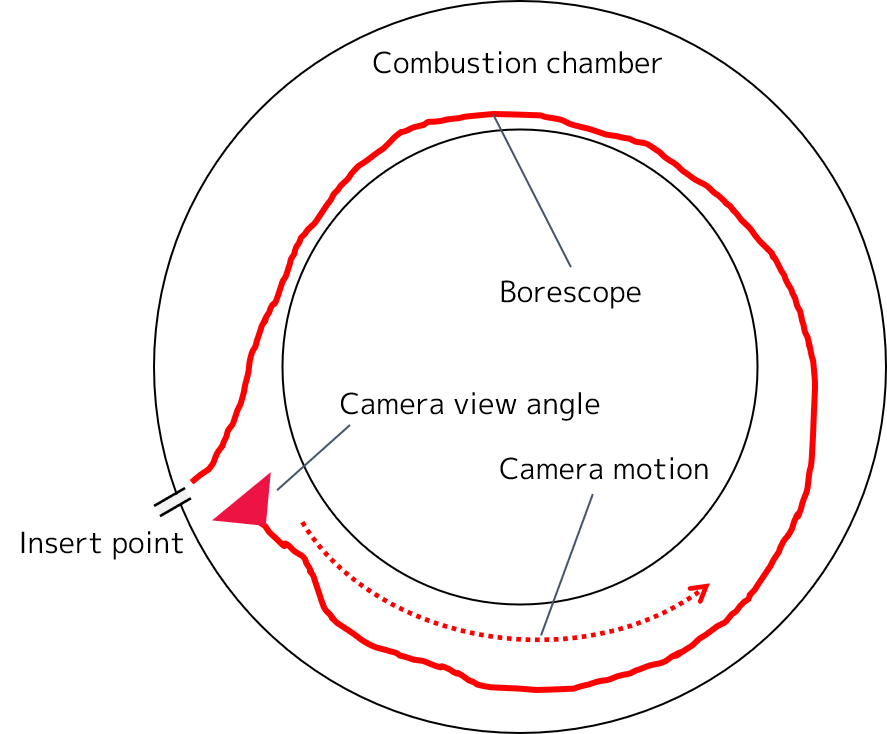}
    \caption{A rough sketch of the inspection for a combustion chamber. }
    \label{fig:inspection}
\end{figure}
\begin{figure}[t]
    \centering
    \includegraphics[width=0.95\linewidth]{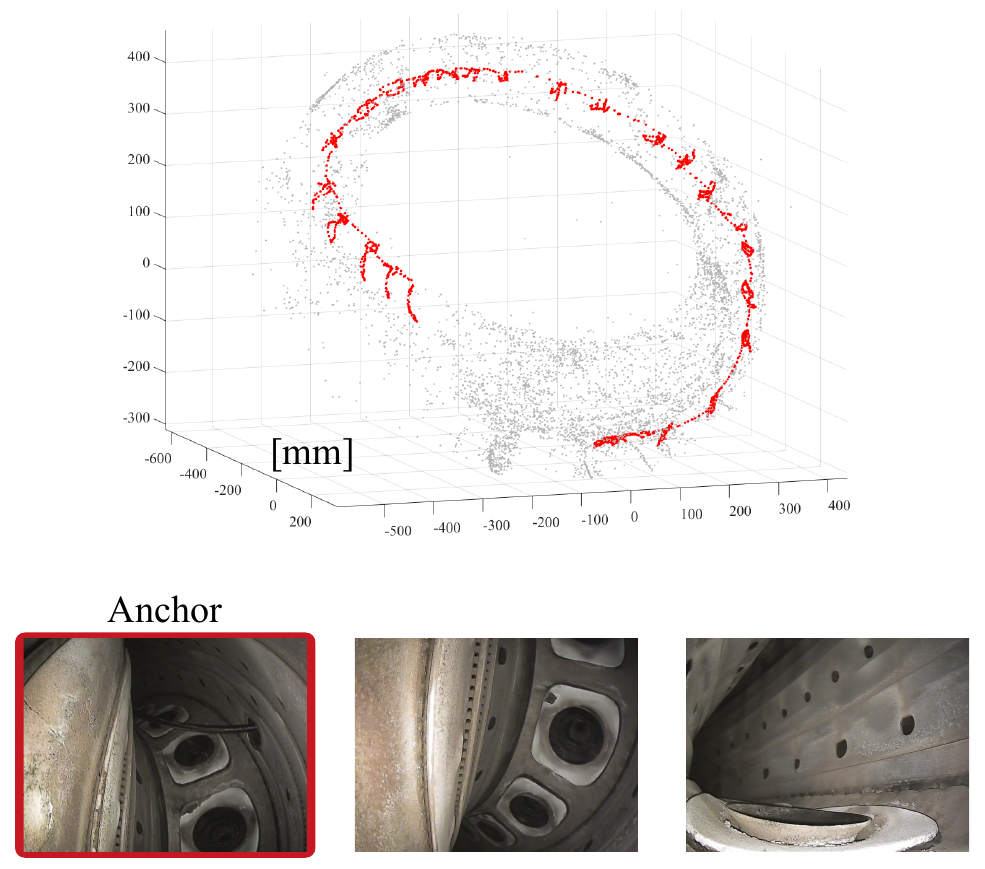}
    \caption{{\bf The test environment and image samples}. Top: Gray dots represent the 3D reference model of the combustion chamber in a jet engine, while red dots are the ground-truth camera location of the test sequence. Bottom: Video frames in the test sequence. \label{fig:testscene}}
\end{figure}
%
%
\renewcommand{\thisheight}{0.24\linewidth}
{\tabcolsep=1pt
\begin{figure*}
    \centering
    {\footnotesize
    \begin{tabular}{cccl}
    \includegraphics[height=\thisheight]{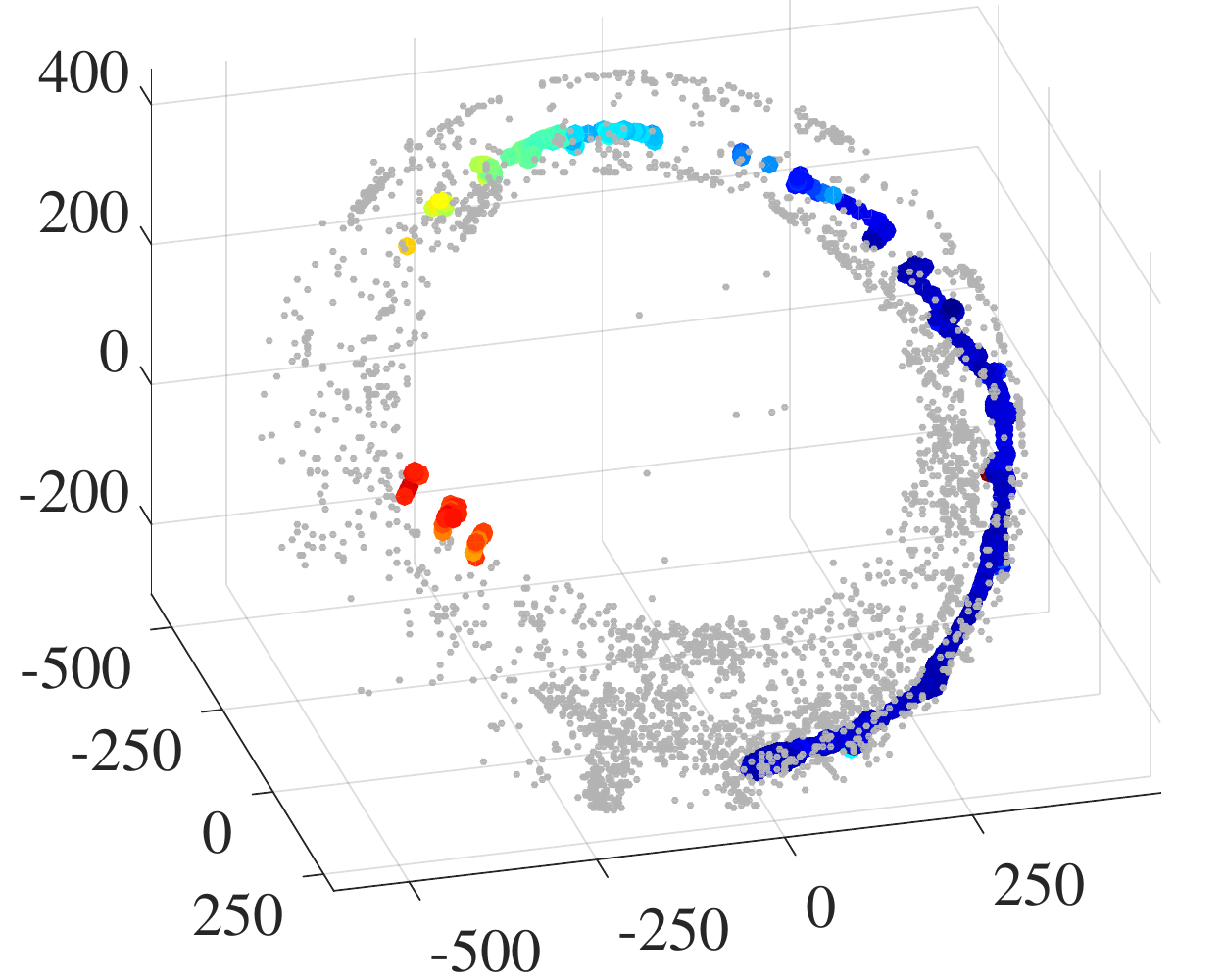} & 
    \includegraphics[height=\thisheight]{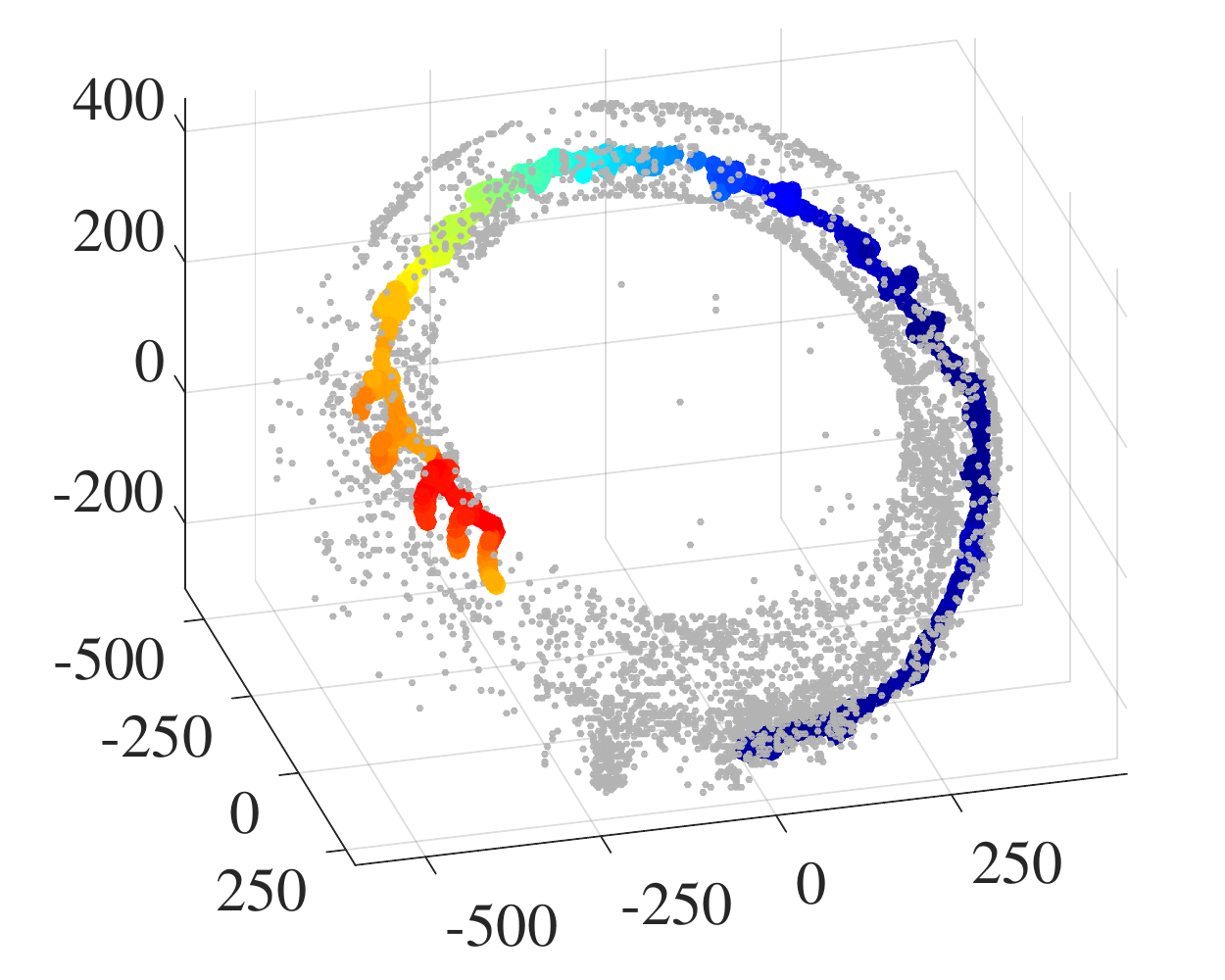} & 
    \includegraphics[height=\thisheight]{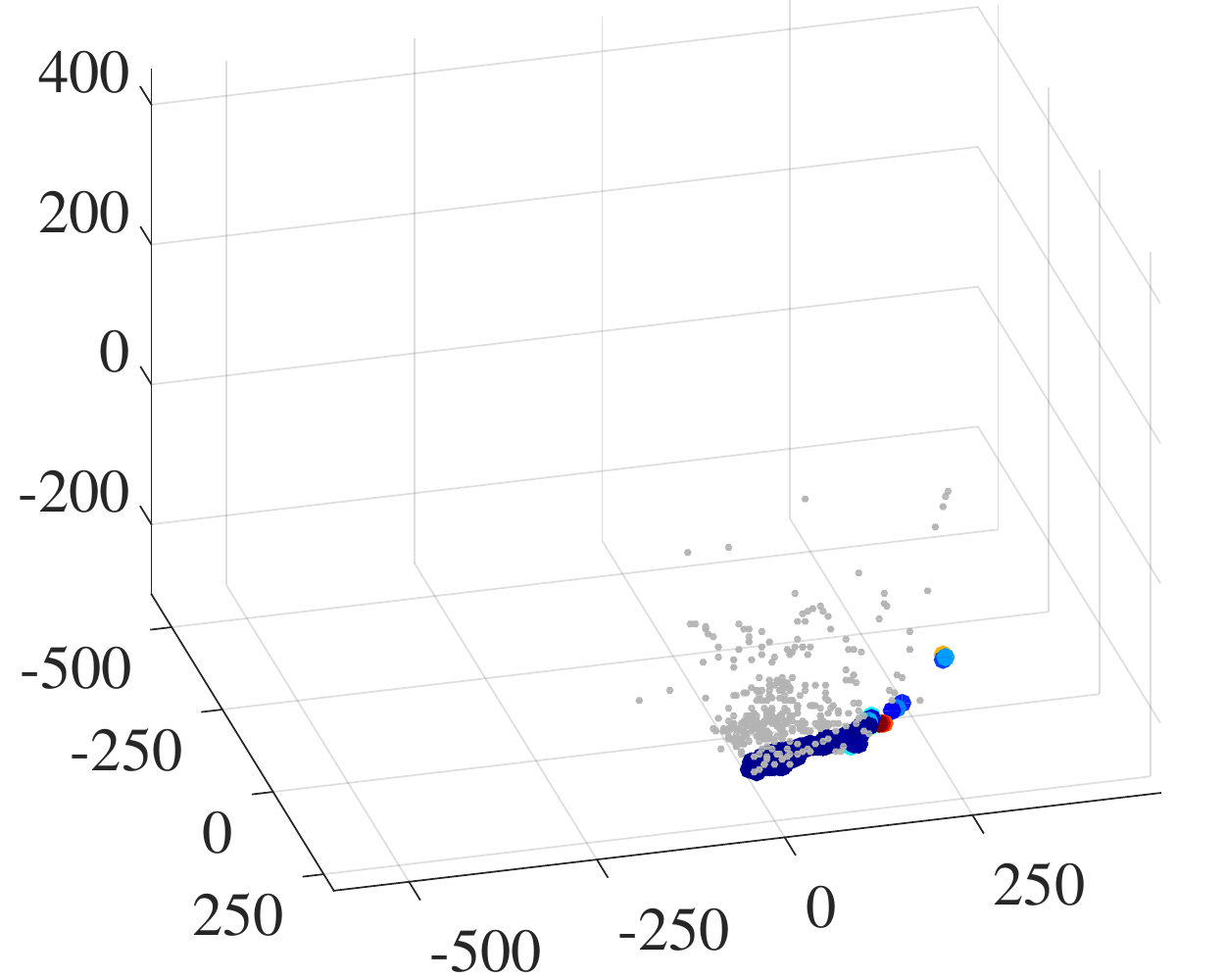} & 
    \includegraphics[height=\thisheight]{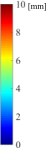} \\[2pt]
    (a) Single image (1,450 cameras) & (b) {\bf Ours (2,247 cameras)} & (c) On-the-fly SfM (501 cameras) &  \\[-1pt]
    \end{tabular}
    }
    \caption{{\bf Reconstructed camera trajectory}. Each camera is colored by its positional error (mm). \label{fig:resultplots}}
\end{figure*}
}
{\tabcolsep=3pt
\begin{table}[t]
    \centering
    \caption{Localization results. \label{tab:stats}}
    \begin{tabular}{r|c|c|c}
    & & & \\[-5pt]
    Method & \#Cameras & MAE & Median error \\ 
     & & [mm] & [mm] \\ \hline
    Single image & 1,450 (63.6\%) & 1.34 & 0.60 \\
    {\bf Ours} & {\bf 2,247 (98.6\%)} & 2.44 & 0.42 \\
    \end{tabular}
\end{table}
}
\section{Experiments \label{sec:exp}}
\para{Test scene: The inside of a jet engine}. We test our localization system during the inspection for a combustion chamber in an aircraft jet engine~\cite{cfmengine}. 
The inspection is usually done by inserting an industrial borescope through an insert point as to go around the chamber (\cf Fig.~\ref{fig:inspection}). 
Then the inspector observes the inside of the chamber via a monocular camera of the borescope, while pulling out the borescope. 
We capture three image sequences during independent trials, two as the database sequences for constructing the initial SfM model, and the other as the query sequences for testing. 
The reference model is constructed by COLMAP~\cite{Schonberger-CVPR16}, resulting in a model consisting of 
5,107 cameras and 1.5M scene points. 
Separately, we gather the ground-truth cameras of query sequences by constructing another SfM model for the query and one of the database sequence. The reconstructed query cameras are registered to the reference model by estimating the similarity transform between the shared frames~\cite{umeyama1991least}, resulting in 2,254 query frames annotated with the ground-truth location, out of 2,280 query frames (\cf Fig.~\ref{fig:testscene}). 

\para{CNN anchor detector}. According to the trajectory of the borescope illustrated in Fig.~\ref{fig:inspection}, the beginning of video often captures the insert point (and the inserted borescope itself), which has a special appearance compared to other part of the chamber (\cf Fig.~\ref{fig:testscene}). 
Therefore we build a CNN classifier to detect such unique frames as anchors. 
The model is based on ResNet18~\cite{he2016deep} architecture, while discarding last two layers (conv4 and conv5, namely) and modifying the final fully connected layer to obtain an one dimensional score of the input image to be an anchor. 
We manually annotate 518 database images (roughly 10\% of all frames) which see the insert point and train the model by minimizing a standard margin loss~\cite{Arandjelovic18NetVLAD}. 
In the testing phase, we feed each of input frames to the trained model, and gather a subset of 32 anchor frames. 

\para{Implementation}. We implement our method mainly based on COLMAP~\cite{Schonberger-CVPR16}, a well-known SfM tool. 
We modify bundle adjustment~\cite{ceres-solver} so that we can freeze the reference model. 
The anchor detector is implemented using PyTorch library. 

\para{Results}. As the main comparison opponent to our method, we also evaluate a baseline localization method for single image: For each input frame, we gather 20 similar database images via image retrieval and match their local features. Using these correspondences, the camera pose of each frame is independently estimated by solving PnP. 

Tab.~\ref{tab:stats} reports the statistics of the localized cameras. Our method can localize further more cameras than the single image method, while remaining the accuracy. 
Fig.~\ref{fig:resultplots} also plots the localized cameras in the model. Due to the challenging scene nature and partly incomplete reconstruction of the reference model, single image method (a) fails to find locations of the latter part of the input sequence. On the other hand, our method (b) achieves a continuous camera trajectory within acceptable errors, which prove the dominance of our approach using multiple sequential images to support each other location. 

As an alternative approach which performs {\it on-the-fly} scene reconstruction~\cite{sattler2017large,torii2019large}, we also construct an SfM model via standard incremental SfM implemented by COLMAP~\cite{Schonberger-CVPR16}, using only input frames. The model is then registered to the scene by estimating the similarity transform between reconstructed cameras and their ground-truth location. 
This approach, however, reconstructs only part of the scene seen from few images (Fig.~\ref{fig:resultplots}~(c)). 
This result clearly points the fact that the well known camera tracking approach that simultaneously estimates 3D structure and camera locations, including ``popular'' SLAM approaches, actually cannot deal with severe appearances such as in industrial scenes, whereas our method can still obtain an accurate camera trajectory while exploiting the pre-constructed 3D model.

\section{Conclusion}
\noindent
In this paper, we have proposed a new camera localization system designed for an image sequence captured in the challenging industrial scene. 
Our method starts reconstruction from a reliable anchor frame that captures an unique object in the scene, and sequentially register neighbor frames while exploiting recently registered frames as the location prior for the new frame. 
In the experiment on an industrial parts inspection scenario, the proposed method achieves an accurate and stable camera trajectory whereas other methods can localize only a part of the sequence. 
We believe our recursive 3D reconstruction anchored to any static object in the scene, is also beneficial for localization problems in many robotics situations, where the operator can perceive the current location and the surrounding environment through the augmented 3D model obtained during the reconstruction.  
One of the future work would be to achieve a real-time processing for a sequential image stream from a camera, so that the system can provide the augmented model in a practical timing. 

\para{Acknowledgement}. This work is partly supported by JSPS KAKENHI Grant Number 17H00744.

\bibliographystyle{IEEEtran}
\bibliography{bib/shortstrings,bib/taira}
\end{document}